\documentclass[11pt,a4paper]{article}
\usepackage{times,latexsym}
\usepackage{url}
\usepackage[T1]{fontenc}

\usepackage[acceptedWithA]{tacl2021v1}

\usepackage{xspace,mfirstuc,tabulary}

\newif\iftaclinstructions
\taclinstructionsfalse 
\iftaclinstructions
\renewcommand{\confidential}{}
\renewcommand{\anonsubtext}{(No author info supplied here, for consistency with
TACL-submission anonymization requirements)}
\newcommand{\instr}
\fi

\iftaclpubformat 

\else

\fi

\usepackage{times}
\usepackage{latexsym}

\usepackage{amsmath}
\usepackage{amsfonts}
\usepackage{booktabs}
\usepackage{graphicx}
\usepackage{xargs}
\usepackage{algorithm}
\usepackage{algpseudocode}
\usepackage{float}
\usepackage{bm}

\usepackage{bbm}
\usepackage{comment}
\usepackage{todonotes}
\usepackage{svg}
\usepackage{mathtools}
\usepackage{enumitem}

\usepackage{array,etoolbox}
\newcolumntype{H}{>{\setbox0=\hbox\bgroup}c<{\egroup}@{}}  
\preto\tabular{\setcounter{magicrownumbers}{0}}
\newcounter{magicrownumbers}

\usepackage[english]{babel}
\usepackage{amsthm}

\newcommand{\R}{\mathbb{R}}

\newcommand{\E}{\mathbb{E}}
\newcommand{\X}{\mathcal{X}}
\newcommand{\Y}{\mathcal{Y}}

\newcommand{\D}{\mathcal{D}}

\newcommand{\n}{{1:n}}

\newcommand\given[1][]{\:#1\vert\:}

\newcommand{\words}{x_\n}
\newcommand{\sent}{\words}

\newcommand{\pos}{t}
\newcommand{\poss}{\pos_\n}

\newcommand{\arc}{e}
\newcommand{\arcs}{\arc_\n}

\newcommand{\dep}{r}
\newcommand{\deps}{\dep_\n}

\newcommand{\Poss}{\mathcal{T}}

\newcommand{\Deps}{\mathcal{R}}

\newcommand{\mjc}[1]{}

\title{Improving Low-Resource Cross-lingual Parsing\\with Expected Statistic Regularization}

\author{
  Thomas Effland\\
  Columbia University\\
  {\tt teffland@cs.columbia.edu}
  \And
  Michael Collins\\
  Google Research\\
  {\tt mjcollins@google.com}
}

\date{}

\begin{document}
\maketitle
\begin{abstract}
We present Expected Statistic Regularization (ESR), a novel regularization technique that utilizes low-order multi-task structural statistics to shape model distributions for semi-supervised learning on low-resource datasets. We study ESR in the context of cross-lingual transfer for syntactic analysis (POS tagging and labeled dependency parsing) and present several classes of low-order statistic functions that bear on model behavior. Experimentally, we evaluate the proposed statistics with ESR for unsupervised transfer on 5 diverse target languages and show that all statistics, when estimated accurately, yield improvements to both POS and LAS, with the best statistic improving POS by +7.0 and LAS by +8.5 on average. We also present semi-supervised transfer and learning curve experiments that show ESR provides significant gains over strong cross-lingual-transfer-plus-fine-tuning baselines for modest amounts of label data. These results indicate that ESR is a promising and complementary approach to model-transfer approaches for cross-lingual parsing.\footnote{We have published for our implementation and experiments at \url{https://github.com/teffland/expected-statistic-regularization}.}
\end{abstract}

\section{Introduction}


In recent years, great strides have been made on linguistic analysis for low-resource languages. These gains are largely attributable to transfer approaches from (1) massive pretrained multilingual language model (PLM) encoders~\citep{devlin2018bert,liu2019roberta}; (2) multi-task training across related syntactic analysis tasks~\citep{Kondratyuk201975L1}; and (3) multi-lingual training on diverse high-resource languages~\citep{wu-dredze-2019-beto,ahmad-etal-2019-cross,Kondratyuk201975L1}. Combined, these approaches have been shown to be particularly effective for cross-lingual syntactic analysis, as shown by UDify~\citep{Kondratyuk201975L1}.

However, even with the improvements brought about by these techniques, transferred models still make syntactically implausible predictions on low-resource languages, and these error rates increase dramatically as the target languages become more distant from the source languages~\citep{He2019CrossLingualST,Meng2019TargetLC}. In particular, transferred models often fail to match many low-order statistics concerning the typology of the task structures. We hypothesize that enforcing regularity with respect to estimates of these structural statistics -- effectively using them as weak supervision -- is complementary to current transfer approaches for low-resource cross-lingual parsing. 

To this end, we introduce Expected Statistic Regularization (ESR), a novel differentiable loss that regularizes models on unlabeled target datasets by minimizing deviation of descriptive statistics of model behavior from target values. The class of descriptive statistics usable by ESR are expressive and powerful. 
For example, they may describe cross-task interactions, encouraging the model to obey structural patterns that are not explicitly tractable in the model factorization.
Additionally, the statistics may be derived from constraints dictated by the task formalism itself (such as ruling out invalid substructures) or by numerical parameters that are specific to the target dataset distribution (such as relative substructure frequencies).
In the latter case, we also contribute a method for selecting those parameters using small amounts of labeled data, based on the bootstrap~\citep{Efron1979BootstrapMA}.

Although ESR is applicable to a variety of problems, we study it using modern cross-lingual syntactic analysis on the Universal Dependencies data, building off of the strong model-transfer framework of UDify~\citep{Kondratyuk201975L1}. We show that ESR is complementary to transfer-based approaches for building parsers on low-resource languages. We present several interesting classes of statistics for the tasks and perform extensive experiments in both oracle unsupervised and realistic semi-supervised cross-lingual multi-task parsing scenarios, with particularly encouraging results that significantly outperform state-of-the-art approaches for semi-supervised scenarios. We also present ablations that justify key design choices.

\section{Expected Statistic Regularization}
\label{s:general}

We consider structured prediction in an abstract setting
where we have inputs $x \in \X$, output structures $y \in \Y$, and a conditional model $p_\theta(y|x) \in \mathbb{P}$ with parameters $\theta \in \Theta$, where $\mathbb{P}$ is the distribution space and $\Theta$ is the parameter space. In this section we assume that the setting is semi-supervised, with a small labeled dataset $\D_L$ and a large unlabeled dataset $\D_U$; let $\D_L = \{ (x_i,y_i)\}_{i=1}^{m}$ be the labeled dataset of size $m$ and similarly define $\D_U = \{ x_i\}_{i=m+1}^{m+n}$ as the unlabeled dataset.

Our approach centers around a vectorized statistic function $f$ that maps unlabeled samples and models to real vectors of dimension $d_f$:
\begin{equation}
    f : \mathbb{D} \times \mathbb{P} \rightarrow \R^{d_f}
\end{equation} 
where $\mathbb{D}$ is the set of unlabeled datasets of any size, (i.e., $\D_U \in \mathbb{D}$).
The purpose of $f$ is to summarize various properties of the model using the sample. For example, if the task is part-of-speech tagging, one possible component of $f$ could be the expected proportion of {\tt NOUN} tags in the unlabeled data $\D_U$. 
In addition to $f$, we assume that we are given vectors of target statistics  $t  \in \R^d$ and margins of uncertainty $\sigma  \in \R^d$ as its supervision signal. We will discuss the details of $f$, $t$, and $\sigma$ shortly but first describe the overall objective. 

\subsection{Semi-Supervised Objective}

Given labeled and unlabeled data $\D_L$ and $\D_U$, we propose the following semi-supervised objective $O$, which breaks down into a sum of supervised and unsupervised terms $L$ and $C$:
\begin{gather}
    \hat{\theta} = \arg\min\limits_{\theta \in \Theta} O(\theta;\D_L,\D_U)\\
    O(\theta;\D_L, \D_U) = L(\theta;\D_L) + \alpha C(\theta;\D_U) \label{eq:O}
\end{gather}
where $\alpha > 0$ is a balancing coefficient. 
The supervised objective $L$ can be any suitable supervised loss; here we will use the negative log-likelihood of the data under the model. Our contribution is the unsupervised objective $C$. 

For $C$, we propose to minimize some distance function $\ell$ between the target statistics $t$ and the value of the  statistics $f$ calculated using unlabeled data and the model $p_\theta$. ($\ell$ will also take into account the uncertainty margins $\sigma$.) A simple objective would be:
\begin{equation*}
    C(\theta; \D_U) = \ell(t, \sigma, f(\D_U, p_\theta))
\end{equation*}
This is a dataset-level loss penalizing divergences from the target level statistics. The problem with this approach is that this is not amenable to modern hardware constraints requiring SGD. Instead, we propose to optimize this loss in expectation over unlabeled mini-batch samples $\D^k_U$, where $k$ is the mini-batch size and $\D^k_U$ is sampled uniformly with replacement from $\D_U$. Then, $C$ is given by:
\begin{equation}
    \label{eq:C}
    C(\theta; \D_U) = \E_{\D^k_U}[ \ell(t, \sigma, f(\D^k_U, p_\theta)) ]
\end{equation}


This objective penalizes the model if the statistic $f$, when applied to samples of unlabeled data $\D^k_U$, deviates from the targets $t$ and thus pushes the model toward satisfying these target statistics.


Importantly, the objective in Eq.~\ref{eq:C} is more general than typical objectives in that the outer loss function $\ell$ does not necessarily break down into a sum over individual input examples---the aggregation over examples is done inside $f$:
\begin{equation}
    \label{eq:not_emr}
    \ell(t, \sigma, f(\D_U, p_\theta))  \neq \sum\limits_{x \in \D_U} \ell(t, \sigma, f(x, p_\theta))
\end{equation}
This generality is useful because components of $f$ may describe statistics that aggregate over inputs, estimating expected quantities concerning sample-level regularities of the structures. In contrast, the right-hand side of Eq.~\ref{eq:not_emr} is more stringent, imposing that the statistic be the same for all instances of $x$. In practice, this loss reduces noise compared to a per-sentence loss, as is shown in Section~\ref{s:batch-ablation}.



\subsection{The Statistic Function $f$}
\label{s:f-details}

In principle the vectorized statistic function $f$ could be almost any function of the unlabeled data and model, provided it is possible to obtain its gradients w.r.t. the model parameters $\theta$,
however, in this work we will assume $f$ has the following three-layer structure.

First, let $g$ be another vectorized function of "sub-statistics" that may have a different dimensionality than $f$ and takes individual $x,y$ pairs as input:
\begin{equation}
    g : \X \times \Y \rightarrow \R^{d_g}
\end{equation} 
Then let $\bar{g}$ be the expected value of $g$ under the model $p_\theta$ summed over the sample $\D_U$:
\begin{equation}
    \label{eq:barg}
    \bar{g} = \sum\limits_{x \in \D_U} \E_{p_\theta(y|x)}[ g(x,y)]
\end{equation}
Given $\bar{g}$, let the $f$'s $j$'th component be the result of an aggregating function $h_j : \R^{d_g} \rightarrow \R$ on $\bar{g}$:
\begin{equation}
    f_j(\D_U, p_\theta) = h_j(\bar{g})
\end{equation}

The individual components $g_i$ will mostly be counting functions that tally various substructures in the data. The $\bar{g}_i$'s then are expected substructure counts in the sample, and the $h_j$'s aggregate small subsets of these intermediate counts in different ways to compute various marginal probabilities. Again, in general $f$ does not need to follow this structure and any suitable statistic function can be incorporated into the regularization term proposed in Eq.~\ref{eq:C}. 

In some cases---when the structure of $g$ does not follow the model factorization either additively or multipicatively---computation of the model expectation $\E_{p_\theta(y|x)}[ g(x,y)]$ in Eq.~\ref{eq:barg} is intractable. In these situations, standard Monte Carlo approximation breaks differentiability of the objective w.r.t. the model parameters $\theta$ and cannot be used. To remedy this, we propose to use the ``Stochastic Softmax'' differentiable sampling approximation from \citet{Paulus2020GradientEW} to allow optimization of these functions. We propose several such statistics in the application (see Section~\ref{s:stats}).

\subsection{The Distance Function $\ell$}
For the distance function $\ell$, we propose to use a smoothed hinge loss~\citep{girshick2015fast} that adapts with the margins $\sigma$. Letting $\bar{f} = f(\D^k_U, p_\theta)$, 
 the $i$'th component of $\ell$ is given by:
\begin{equation}
  \label{eq:smooth-l1}
  \ell_i = \begin{cases}
    \frac{(\bar{f}_i - t_i)^2}{2 \sigma_i} & \text{if}\ \ |\bar{f}_i- t_i | < \sigma_i\\
    | \bar{f}_i - t_i| -  \sigma_i & \text{else}\ \ \\
    \end{cases}
\end{equation}
The total loss $\ell$ is then the sum of its components:
\begin{equation}
    \ell(t, \sigma, f(\D^k_U, p_\theta)) = \sum_i \ell_i(t_i, \sigma_i, \bar{f}_i)
\end{equation}

We choose this function because it is robust to outliers, adapts its width to the margin parameter $\sigma_i$, and expresses a preference for $f_i = t_i$ (as opposed to max-margin losses). We give an ablation study in Section~\ref{s:loss-ablation} justifying its use.


\section{Choosing the Targets and Margins}
\label{s:targets}
There are several possible approaches to choosing the targets $t$ and margins $\sigma$, and in general they can differ based on the individual statistics. For some statistics it may be possible to specify the targets and margins using prior knowledge or formal constraints from the task. 
In other cases, estimating the targets and margins may be more difficult. Depending on the problem context, one may be able to estimate them from related tasks or domains (such as neighboring languages for cross-lingual parsing). Here, we propose a general method that estimates the statistics using labeled data, and is applicable to semi-supervised scenarios where at least a small amount of labeled data is available.


The ideal targets are the expected statistics under the ``true'' model $p^*$ are: $t^* =\E_{\D^k_U}[f(\D^k_U, p^*)]$, where $k$ is the batch size. We can estimate this expectation using labeled data $\D_L$ and bootstrap sampling~\citep{Efron1979BootstrapMA}. Utilizing $\D_L$ as a set of point estimates for $p^*$, we sample $B$ total minibatches of $k$ labeled examples uniformly with replacement from $\D_L$ and calculate the statistic $f$ for each of these minibatch datasets. We then compute the target statistic as the sample mean:
\begin{equation}
    \label{eq:bootstrap-mean}
    t = \frac{1}{B}\sum_{i=1}^B f(\D^{(i)}_L)\ ,\ \ \   |\D^{(i)}_L| = k,\ \forall i 
\end{equation}
where we have slightly abused notation by writing $f(\D_L)$ to mean $f$ computed using the inputs $\{ x : (x,y) \in \D_L \}$ and the point estimates $p^*(y|x) = 1,\ \forall (x,y) \in \D_L$. 

In addition to estimating the target statistics for small batch sizes, the bootstrap gives us a way to estimate the natural variation of the statistics for small sample sizes.  To this end, we propose to utilize the standard deviations from the bootstrap samples as our margins of uncertainty $\sigma$:
\begin{equation}
    \label{eq:bootstrap-std}
    \sigma = \sqrt{\frac{1}{B-1} \sum\limits_{i=1}^B (f(\D_L^{(i)}) - t)^2}
\end{equation}
This allows our loss function $\ell$ to adapt to more or less certain statistics. If some statistics are naturally too variable to serve as effective supervision, they will automatically have weak contribution to $\ell$ and little impact on the model training.




\section{Application to Cross-Lingual Parsing}

Now that we have described our general approach, in this section we lay out a proposal for applying it to cross-lingual joint POS tagging and dependency parsing. We choose to apply our method to this problem because it is an ideal testbed for controlled experiments in semi-supervised structured prediction. By their nature, the parsing tasks admit many types of interesting statistics that capture cross-task, universal, and language-specific facts about the target test distributions. 

We evaluate in two different transfer settings: oracle unsupervised and realistic semi-supervised. In the oracle unsupervised settings, there is no supervised training data available for the target languages (and the $L$ term is dropped from Eq.~\ref{eq:O}), but we use target values and margins calculated from the held-out supervised data. This setting allows us to understand the impact of our regularizer in isolation without the confounding effects of direct supervision or inaccuracte targets. In the semi-supervised experiments, we vary the amounts of supervised data, and calculate the targets from the small supervised data samples. This is a realistic application of our approach that may be applied to low-resource learning scenarios.


\subsection{Problem Setup and Data}
\label{s:data}


We use the Universal Dependencies~\citep{Nivre2020MultilingualDP} v2.8 (UD) corpus as data. In UD, syntactic annotation is formulated as a labeled bilexical dependency tree, connecting words in a sentence, with additional part-of-speech (POS) \textit{tags} annotated for each word. The labeled tree can be broken down into two parts: the \textit{arcs} that connect the \textit{head} words to \textit{child} words, forming a tree, and the dependency \textit{labels} assigned to each of those arcs. 
Due to the definition of UD syntax, each word is the child of exactly one arc, and so both the attachments and labels can be written as sequences that align with the words in the sentence.

More formally then, for each labeled sentence $\sent$ of length $n$, the full structure $y$ is given by the three sequences $y = (\poss, \arcs, \deps)$, where $\poss,\ \pos_i \in \Poss$ are the POS tags, $\arcs,\ \arc_i \in \{1,\dots,n\}$ are the head attachments, and $\deps,\ \dep_i \in \Deps$ are the dependency labels.



\subsection{The Model and Training}
\label{s:model}

We now turn to the parsing model that is used as the basis for our approach. Though the general ideas of our approach are adaptable to other models, we choose to use the  UDify architecture because it is one of the state-of-the-art multilingual parsers for UD. 

\subsubsection{The UDify Model}

The UDify model is based on trends in state-of-the-art parsing, combining a multilingual pretrained transformer language model encoder (mBERT) with a deep biaffine arc-factored parsing decoder, following \citet{Dozat2017DeepBA}. These encodings are additionally used to predict POS tags with a separate decoder. 
The full details are given in \citet{Kondratyuk201975L1}, but here it suffices to characterize the parser by its top-level probabilistic factorization:
\begin{align}
    &p(\poss, \arcs, \deps|\sent) \nonumber\\
    &= p(\arcs|\sent) p(\poss|\sent)p(\deps|\arcs,\sent)\\
    &= p(\arcs|\sent) \prod\limits_{i=1}^n p(\pos_i|\sent) p(\dep_i|\arc_i,\sent)
\end{align}

This model is scant on explicit joint factors, following recent trends in structured prediction that forgo higher-arity factors, instead opting for shared underlying contextual representations produced by a mBERT that implicitly contain information about the sentence and structure as a whole. This factorization will prove useful in Section~\ref{s:stats} where it will allow us to compute many of the supervision statistics under the model exactly.


\subsubsection{Training}
\label{s:training}

The UDify approach to training is simple: it begins with a multilingual PLM, mBERT, then fine-tunes the parsing architecture on the concatenation of the source languages. With vanilla UDify, transfer to target languages is zero-shot. 

Our approach begins with these two training steps from UDify, then adds a third: adapting to the target language using the target statistics and possibly small amounts of supervised data (Eq.~\ref{eq:O}).



\subsection{Typological Statistics as Supervision}
\label{s:stats}

We now discuss a series of statistics that we will use as weak supervision. Most of the proposed statistics describe various probabilities for different (but related) grammatical substructures and can ultimately be broken down into ratios of ``count'' functions (sums of indicators), which tally various types of events in the data. 
We propose statistics that cover surface level (POS-only), single-arc, two-arc, and single-head substructures, as well as conditional variants. Due to space constraints, we omit their mathematical descriptions.

\vspace{0.25em}
\noindent \textbf{Surface Level:} One simple set of descriptive statistics are the unigram and bigram distributions over POS tags. POS unigrams can capture some basic relative frequencies, such as our expectation that nouns and verbs are common to all languages.
POS bigrams will allow us to capture simple word-order preferences. 

\noindent \textbf{Single-Arc:} This next set of statistical families all capture information about various choices in single-arc substructures.  A single arc substructure carries up to 5 pieces of information: the arc's direction, label, and distance, as well as the tags for the head and child words. Various subsets of these capture differing forms of regularity, such as ``the probability of seeing tag $\pos_h$ head an arc with label $\dep$ in direction $d$. 

\noindent \textbf{Universally Impossible Arcs:} In addition to many single-arc variants, we also consider the specific subset of (head tag, label, child tag) single-arc triples that are never seen in the any UD data. These combinations, correspond to the impossible arrangements that do not ``type-check'' within the UD formalism and are interesting in that they could in principle be specified by a linguist without any labeled data whatsoever. 
As such, they represent a particularly attractive use-case of our approach, where a domain expert could rule out all invalid substructures  dictated from the task formalism without the model having to learn it implicitly from the training data. With complex structures, this can be a large proportion of the possibilities: in UD we can rule out 93.2\% (9,966/10,693) of the combinations.

\noindent \textbf{Two-Arc:} We also consider substructures spanning two connected arcs in the tree. They may be useful because they cover many important typological phenomena, such as subject-object-verb ordering. They also have been known to be strong features in higher-order parsing models, such as the parser of \citet{carreras-2007-experiments}, but are also known to be intractable in non-projective parsers~\citep{McDonald2006OnlineLO}.

Following \citet{McDonald2006OnlineLO}, we distinguish between two different patterns of neighboring arcs: \textit{siblings} and \textit{grandchildren}. Sibling arc pairs consist of two arcs which share a single head word, while grandchild arc pairs share an intermediate word that is the child of one arc and the head of another. 

\noindent \textbf{Head-Valency:} One interesting statistic that does not fall into the other categories is the valency of a particular head tag. This corresponds to the count of outgoing arcs headed by some tag. We convert this into a probability by using a binning function that allows us to quantify the ``probability that some tag heads between $a$ and $b$ children''. Like the two-arc statistics, expected valency statistics are intractable under the model and we must approximate their computation. 

\noindent \textbf{Conditional Variants:}
Further, each of these statistics can be described in conditional terms, as opposed to their full joint realizations. To do this, we simply divide the joint counts by the counts of the conditioned-upon sub-events. 
Conditional variants may be useful because they do not express preferences for probabilities of the sub-events on the right side of the conditioning bar, which may be hard to estimate. 


\noindent \textbf{Average Entropy:} In addition to the above proposed relative frequency statistics, we also include average per-token, per-edge, and MST tree entropies as additional regularization statistics that are always used. Though we do not show it here, each of these functions may be formulated as a statistic within our approach. The inclusion of these statistics amounts to a form of Entropy Regularization~\citep{Grandvalet2004SemisupervisedLB} that keep the models from optimizing the other ESR constraints with degenerate constant predictions~\citep{Mann2010GeneralizedEC}.



\section{Oracle Unsupervised Experiments}

We begin with oracle unsupervised transfer experiments that evaluate the potential of many types of statistics and some ablations. In this setting, we do not assume any labeled data in the target language, but do assume accurate target statistics and margins, calculated from held-out training data using the method of Section~\ref{s:targets}. This allows us to study the potential of our proposed ESR regularization term $C$ on its own and without the confounds of supervised data or inaccurate targets. 


\subsection{Experimental Setup}

Next we describe setup details for the experiments. These settings additionally apply to the rest of the experiments unless otherwise stated.

\subsubsection{Datasets}

In all experiments, the models are first initialized from mBERT, then trained using the UDify code~\citep{Kondratyuk201975L1} on 13 diverse treebanks, following \citet{Kulmizev2019DeepCW,Ustun2020UDapterLA}. This model, further referred to as \textsc{\textbf{UDpre}}, is used as the foundation for all approaches. 

As discussed in \citet{Kulmizev2019DeepCW}, these 13 training treebanks were selected to give a diverse sample of languages, taking into account factors such as language families, scripts, morphological complexity, and annotation quality.

We evaluate all proposed methods on 5 held-out languages, similarly selected for a diversity in language typologies, but with the additional factor of tranfser performance of the \textsc{\textbf{UDpre}} baseline.\footnote{While we would like to evaluate on as many UD treebanks as possible, budgetary constraints required that we restrict the number of test languages when experimenting with settings that combinatorially vary in other dimensions. We do however experiment with more languages in Section~\ref{s:many-langs}.}

A summary table of these training and evaluation treebanks is given in Table~\ref{tbl:datasets}.

\begin{table*}[h!]
    \centering
    \footnotesize
    \begin{tabular}{cccccc}
    Language & Code & Treebank & Family  & Train Sents & \textsc{UDpre} LAS\\
    \toprule
    Arabic & ar & PADT & Semitic  & 6.1k & 80.5\\
    Basque & eu & BDT & Basque & 5.4k & 77.0\\
    Chinese & zh & GSD & Sino-Tibetan  & 4.0k & 62.3\\
    English & en & EWT & IE, Germanic  & 12.5k & 88.1\\
    Finnish & fi & TDT & Uralic  & 12.2k & 84.4\\
    Hebrew & he & HTB & Semitic  & 5.2k & 80.5\\
    Hindi & hi & HDTB & IE, Indic  & 13.3k & 87.0\\
    Italian & it & ISDT & IE, Romance  & 13.1k & 91.8\\
    Japanese & ja & GSD & Japanese & 7.1k & 73.6\\
    Korean & ko & GSD & Korean & 4.4k & 79.0\\
    Russian & ru & SynTagRus & IE, Slavic & 15.0k$^*$ & 89.1\\
    Swedish & sv & Talbanken & IE, Germanic & 4.3k & 85.7\\
    Turkish & tr & IMST & Turkic & 3.7k & 61.7\\
    \midrule
    German & de & HDT & IE, Germanic & 153.0k & 82.7\\
    Indonesian & id & GSD & Austronesian & 4.5k & 50.4\\
    Maltese & mt & MUDT & Semitic & 1.1k & 20.9\\
    Persian & fa & PerDT & IE, Iranian & 26.2k & 57.0\\
    Vietnamese & vi & VTB & Austro-Asiatic & 1.4k & 48.1\\
    \bottomrule
    \end{tabular} 
    \caption{Training and evaluation treebank details. The final column shows \textsc{UDpre} test set performance after UDify training (evaluation treebank performance is zero-shot).
    ($*$): downsampled to the same 15k sentences as \citet{Ustun2020UDapterLA} to reduce training time and balance the data.}
    \label{tbl:datasets}
\end{table*}


\subsubsection{Apporaches}

We compare our approach to two strong baselines in all experiments, based on recent advances in the literature for cross-lingual parsing. These baselines are implemented in our code so that we may fairly compare them in all of our experiments.

\begin{itemize}[topsep=0.25em,itemsep=0em,leftmargin=*]
    \item \textbf{\textsc{UDpre}}: The first baseline is the UDify~\citep{Kondratyuk201975L1} model-transfer approach. Multilingual model-transfer alone is currently one of the state-of-the-art approaches to cross-lingual parsing and is a strong baseline in its own right.
    \item \textbf{\textsc{UDpre-PPT}:} We also apply the Parsimonious Parser Transfer (PPT) approach from \citet{Kurniawan2021PPTPP}. PPT is a nuanced self-training approach, extending \citet{Tckstrm2013TargetLA}, that encourages the model to concentrate its mass on its most likely predicted parses for the target treebank. We use their loss implementation, but apply it to our \textsc{UDpre} base model (instead of their weaker base model) for a fair comparison, so this approach combines UDify with PPT.
    \item \textbf{\textsc{UDpre-ESR}:} Our proposed approach, Expected Statistic Regularization (ESR), applied to \textsc{UDpre} as an unsupervised-only objective. In individual experiments we will specify the statistics used for regularization.
\end{itemize}

\subsubsection{Training and Evaluation Details}

For metrics, we report accuracy for POS tagging, coarse-grained labeled attachment score (LAS) for dependency trees, and their average as a single summary score. The metrics are computed using the official CoNLL-18 evaluation script.\footnote{\url{https://universaldependencies.org/conll18/evaluation.html}} 
For all scenarios, we use early-stopping for model selection, measuring the POS-LAS average on the specified development sets. 

We tune learning rates and $\alpha$ for each proposed loss variant at the beginning of the first experiment with a low-budget grid search, using the settings that achieve best validation metric on average across the 5 language validation sets for all remaining experiments with that variant. We find generally that a base learning rate of $2\times 10^{-5}$ and $\alpha = 0.01$ worked well for all variants of our method. 
We train all models using AdamW~\citep{Loshchilov2019DecoupledWD} on a slanted triangular learning rate schedule~\citep{devlin2018bert} with 500 warmup steps. Also, since the datasets vary in size, we normalize the training schedule to 25 epochs at 1000 steps per epoch. 
We use a batch size of $8$ sentences for training and estimating statistic targets. When bootstrapping estimates for $t$ and $\sigma$ we use $B=1000$ samples.

\subsection{Assessing the Proposed Statistics}
\label{s:stat-variants}

\begin{table*}[t!]
    \centering
    \footnotesize
    \addtolength{\tabcolsep}{-1pt}
    \begin{tabular}{H!{\hspace{0cm}}lrrrrrH!{\hspace{0.175cm}}rrrrrH!{\hspace{0.175cm}}r}
\toprule
& & \multicolumn{6}{c}{POS} & \multicolumn{6}{c}{LAS} & \\
\cmidrule(lr){3-7} \cmidrule(lr){9-13}
& Statistic &   de &  id &  fa  &   vi &    mt &  avg &   de &    id &    fa &   vi &    mt &  avg &  avg \\
\midrule                  
& \textsc{UDpre} & 89.3 & 80.3 & 83.0 & 64.7 & 41.4 & 71.7 & 82.7 & 50.4 & 57.0 & 48.1 & 20.9 & 51.8 &  61.8 \\
& \textsc{UDpre-PPT} & +0.4 & +5.6 & -1.5 & -0.1 & +3.1 & +1.5 & +0.2 & +8.1 & -5.5 & -0.3 & +4.6 & +1.4 & +1.5\\
\midrule
2.&  $^\dagger$\textbf{Child, Label }                         & +3.3 & +5.7 & +8.0 & +4.0 & +14.0 & +7.0     & +3.5 & +10.1 & +18.4 & +0.5 & +10.2 & +8.5 &     +7.8 \\
3.& $^{*\dagger}$Child, Label, Grand-label
           & +1.5 & +2.8 & +5.3 & +5.9 & +15.7 & +6.2 & +2.5 &  +9.2 & +16.2 & +3.2 & +12.5 & +8.7 &     +7.5 \\
4.& $^\dagger$Head, Child, Label                     & +2.5 & +4.9 & +7.2 & +4.1 & +14.2 & +6.6 & +2.8 &  +9.9 & +17.2 & +1.3 & +10.2 & +8.3 &     +7.4 \\
5.& $^\dagger$Head, Label                          & +1.7 & +3.2 & +5.2 & +6.3 & +14.2 & +6.1 & +2.7 &  +9.0 & +16.4 & +3.8 & +11.0 & +8.6 &     +7.3 \\
6.& $^\dagger$Head, Label $\given$ Child          & +3.0 & +5.2 & +5.8 & +5.7 & +10.9 & +6.1 & +2.8 &  +9.7 & +15.3 & +2.1 &  +5.6 & +7.1 &     +6.6 \\
7.& $^\dagger$Label                              & -0.2 & +4.4 & +5.1 & +0.0 & +11.3 & +4.1 & +2.9 &  +8.4 & +17.1 & +4.0 &  +9.5 & +8.4 &     +6.2 \\
8.& $^\dagger$Label, Distance                         & -0.2 & +3.7 & +3.9 & -0.1 & +11.7 & +3.8 & +2.8 &  +8.9 & +16.3 & +4.1 &  +9.4 & +8.3 &     +6.0 \\
9.& $^{*\dagger}$Head, Sibling Children Tags            & +2.2 & +5.1 & +5.6 & +7.8 & +14.0 & +6.9 & +1.6 &  +2.7 & +11.3 & -3.0 & +11.1 & +4.7 &     +5.8 \\
10.& $^\dagger$Head, Child                        & +2.3 & +5.3 & +5.9 & +3.7 & +14.0 & +6.2 & +1.9 &  +3.3 & +12.2 & -0.7 & +10.1 & +5.4 &     +5.8 \\
11.& $^\dagger$Label $\given$ Child             & +1.2 & +4.3 & +4.8 & -0.5 & +10.0 & +4.0 & +3.3 &  +9.5 & +16.6 & +2.0 &  +5.6 & +7.4 &     +5.7 \\
26.& \textbf{Universal Arc}   & +2.4 & +4.7 & +1.4 & +1.4 & +8.7 & +3.7 & +2.1 & +8.1 & +4.0 & -3.1 & +3.9 & +3.0 & +3.4 \\
\bottomrule
\end{tabular}
    \caption{\textit{Unsupervised Oracle Statistic Variant Results.} \textit{(Top):} Baseline methods that do not use ESR. \textit{(Bottom):} Various statistics used by ESR as unsupervised loss on top of \textsc{UDpre}. Scores are measured on target treebank development sets. Bold names mark statistics used in later experiments.
    \textit{($*$):} All statistics with $*$ are intractable and utilize the SST relaxation of \citet{Paulus2020GradientEW}.
    \textit{($\dagger$):} All statistics with $\dagger$ also include directional information.} 
    \label{tbl:all-variants}
\end{table*}

In this experiment we evaluate 32 types of statistics from Section~\ref{s:stats} for transfer of the \textsc{UDpre} model (pretrained on 13 languages) to the target languages. The purpose of this experiment is to get a sense of the effectiveness of each statistic for improving model-based cross-lingual transfer.\footnote{While it would also be possible to try out different combinations of the various statistics, due to cost considerations we leave these experiments to future work.} To prevent overfitting to the test sets for later experiments, all metrics for this experiment are calculated on the development sets.


\noindent \textbf{Results:} The results of the experiment are presented in Table~\ref{tbl:all-variants}, ranked from best to worst. Due to space constraints, we only show the top 10 statistics in addition to the Universal-Arc statistic. Generally we find that all of the 32 proposed statistics improve upon the \textsc{UDpre} and \textsc{UDpre-PPT} models on average, with many exhibiting large boosts. 
The best performing statistic concerns (Child Tag, Label, Direction) substructures, yielding an average improvement of +7.0 POS and +8.5 LAS, an average relative error rate reduction of 23.5\%. Many other statistics are not far behind, and overall statistics that bear on the child tag and dependency label had the highest impact. This indicates that, with accurate target estimates, the proposed statistics are highly complementary to multilingual parser pretraining (\textsc{UDpre}) and substantially improve transfer quality in the unsupervised setting. By comparison, the \textsc{PPT} approach provides marginal gains to \textsc{UDpre} of only +1.4 average POS and +1.5 average LAS.

Another interesting result is that several of the intractable two-arc statistics were among the best statistics overall, indicating that the use of the differentiable SST approximation does not preclude the applicability of intractable statistics. For example the directed grandchild statistic of cooccurrences of incoming and outgoing edges for certain tags was the second highest performing, with an average improvement of +7.0 POS accuracy and +8.5 LAS (21.3\% average error rate reduction).

Results for the conditional variants (not shown) were less positive. Generally, conditional variants were worse than their full joint counterparts (e.g., "Child $\given$ Label" and "Label $\given$ Child" are worse than "Child, Label"), performing worse in 15/16 cases. This makes sense, as we are using accurate statistics and full joints are strictly more expressive. 

This experiment gives a broad but shallow view into the effectiveness of the various proposed statistics. In the rest of the experiments, we  evaluate the following two variants in more depth: 
\begin{enumerate}[topsep=0.25em,itemsep=0em,leftmargin=*]
    \item \textbf{\textsc{ESR-CLD}}, which supervises target proportions for (Child Tag, Label, Direction) triples. This is the ``Child, Label'' row in Table~\ref{tbl:all-variants}.
    \item \textbf{\textsc{ESR-UniArc}}, which supervises the 9,966 universally impossible (Head Tag, Child Tag, Label) arcs that do not require labeled data to estimate. All of these combinations have targets values of $t=0$ and margins $\sigma=0$. This is the ``Universal Arc'' row in Table~\ref{tbl:all-variants}.
\end{enumerate}
We choose these two because \textsc{ESR-CLD} is the best performing statistic overall and \textsc{ESR-UniArc} is unique in that it does not require labeled data to estimate; we do not evaluate others because of cost considerations.

\subsection{Ablation Studies}
\label{s:ablations}

Next, we perform two ablation experiments to evaluate key design choices of the proposed approach. First, we evaluate the use of batch-level aggregation in the statistics before the loss, versus the more standard approach of loss-per-sentence. In the second, we evaluate the proposed form of $\ell$. 

We compare the two aggregation variants using the CLD (Child Tag, Label, Direction) statistic (\textsc{ESR-CLD}). We report test set results averaged over all 5 languages.
We use the same hyperparameters selected in Section~\ref{s:stat-variants}. 

\subsubsection{Batch-level Loss Ablation}
\label{s:batch-ablation}

In this ablation, we evaluate a key feature of our proposal---the aggregation of the statistic over the batch before loss computation Eq.~\ref{eq:not_emr} versus the more standard approach, which is to apply the loss per-sentence. The former, ``Loss per batch'', has the form:
$\ell(t, \sigma, f(\D_U, p_\theta))$
while the latter, ``Loss per sentence'', has the form:
$\sum\limits_{x \in \D_U} \ell(t, \sigma, f(x, p_\theta))$.

The significance of this difference is that ``Loss per batch'' allows for the variation in individual sentences to somewhat average out and hence is less noisy, while ``Loss per sentence'' requires that each sentence individually satisfy the targets.

\noindent \textbf{Results:} The results are presented in Table~\ref{tbl:loss-agg-ablation-mini}. From the table we can see that ``Loss per batch'' has an average POS of 79.9 and average LAS of 60.4, compared to ``Loss per sentence'' with average POS of 77.1 and LAS of 58.5, which amount to +2.8 POS and +1.9 LAS improvements. This indicates that applying the loss at the batch level confers an advantage over applying per sentence.

\begin{table}
    \centering
    \footnotesize
    \begin{tabular}{lccc}
\toprule
Aggregation Variant &  POS avg  & LAS avg &  avg \\
\midrule                  
Loss per sentence  & 77.1 & 58.5 & 67.8 \\
Loss per batch (ESR) & \textbf{79.9} &  \textbf{60.4}  & \textbf{70.1}\\
\bottomrule
\end{tabular}
    \caption{\textit{Loss Aggregation Ablation Results.} Loss per batch outperforms loss per sentence for both POS and LAS on average.} 
    \label{tbl:loss-agg-ablation-mini}
\end{table}

\subsubsection{Smooth Hinge-Loss Ablation}
\label{s:loss-ablation}

Next, we evaluate the efficacy of the proposed smoothed hinge-loss distance function $\ell$. We compare to using just L1 or L2 uninterpolated and with no margin parameters ($\sigma =0$). We also compare to the ``Hard L1'', which is the max-margin hinge $\ell(t,\sigma,x) = \max \{0, |t - x| - \sigma \}$. We use the same experimental setup as the previous ablation.

\noindent \textbf{Results:} The results are presented in Table~\ref{tbl:loss-func-ablation-mini}. From the table we can see that the Smooth L1 loss outperforms the other variants.

\begin{table}
    \centering
    \footnotesize
    \addtolength{\tabcolsep}{-1pt}
    \begin{tabular}{lccc}
\toprule
$\ell$ Variant &   POS avg &   LAS avg &  avg \\
\midrule                  
L2 ($\sigma=0$) & 78.0 & 58.2 & 68.1 \\
L1 ($\sigma=0$) & 78.5 & 60.3 & 69.5 \\
Hard L1 (max-margin) & 78.4 & 59.9 & 69.2 \\
Smooth L1 (ESR) & \textbf{79.9} & \textbf{60.4}  & \textbf{70.1}\\ 
\bottomrule
\end{tabular}
    \caption{\textit{Loss Function Ablation Results.} The Smooth L1 loss outperforms the other simpler loss variants for both POS and LAS, averaged over 5 languages.} 
    \label{tbl:loss-func-ablation-mini}
\end{table}

\section{Realistic Semi-Supervised Experiments}
\label{s:semi-supervised}

The previous experiments considered an unsupervised transfer scenario without labeled data. In these next experiments we turn to a realistic semi-supervised application of our approach where we have access to limited labeled data for the target treebank.

\subsection{Learning Curves}
In this experiment we present learning curves for the approaches, varying the amount of labeled data $|\D^{\text{train}}_L| \in \{ 50, 100, 500, 1000\}$. To make experiments realistic, we calculate the target statistics $t$ and margins $\sigma$ from the small subsampled labeled training datasets using Eqs.~\ref{eq:bootstrap-mean} and \ref{eq:bootstrap-std}.

We study two distinct settings. First, we study the multi-source domain-adaptation transfer setting, \textsc{UDpre}.
Second, we study our approach in a more standard semi-supervised scenario where we cannot utilize intermediate on-task pretraining and domain-adaption, instead learning on the target dataset starting ``from scratch'' with the pretrained PLM (\textsc{mBERT}). 

We use the same baselines as before, but augment each with a supervised fine-tuning loss on the supervised data in addition to any unsupervised losses. We refer to these models as \textbf{\textsc{UDpre-FT}, \textsc{UDpre-FT-PPT}, and \textsc{UDpre-FT-ESR}}. That is, models with \textbf{\textsc{FT}} in the name have some supervised fine-tuning in the target language.

In these experiments, we subsample labeled training data 3 times for each setting. We report averages over all 5 languages, 3 supervised subsample runs each, for a total of 15 runs per method and dataset size.  We also use subsampled development sets so that model selection is more realistic.\footnote{As is argued by \citet{Oliver2018RealisticEO}, using a realistically-sized development set is overlooked in much of the semi-supervised literature, leading to inappropriately strong model selection and overly optimistic results.} For development sets we subsample the data to a size of $|\D_L^{\text{dev}}| = \min(100, |\D^{\text{train}}_L|)$, which reflects a 50/50 train/dev split until $|\D_L| \geq 200$, at which point we maximize training data and only hold out 100 sentences for validation.

We use the same hyperparameters as before, except we use 40 epochs with 200 steps per epoch as the training schedule, mixing supervised and unsupervised data at a rate of 1:4.

\begin{figure*}[t!]
    \centering
    \includegraphics[width=1.0\textwidth]{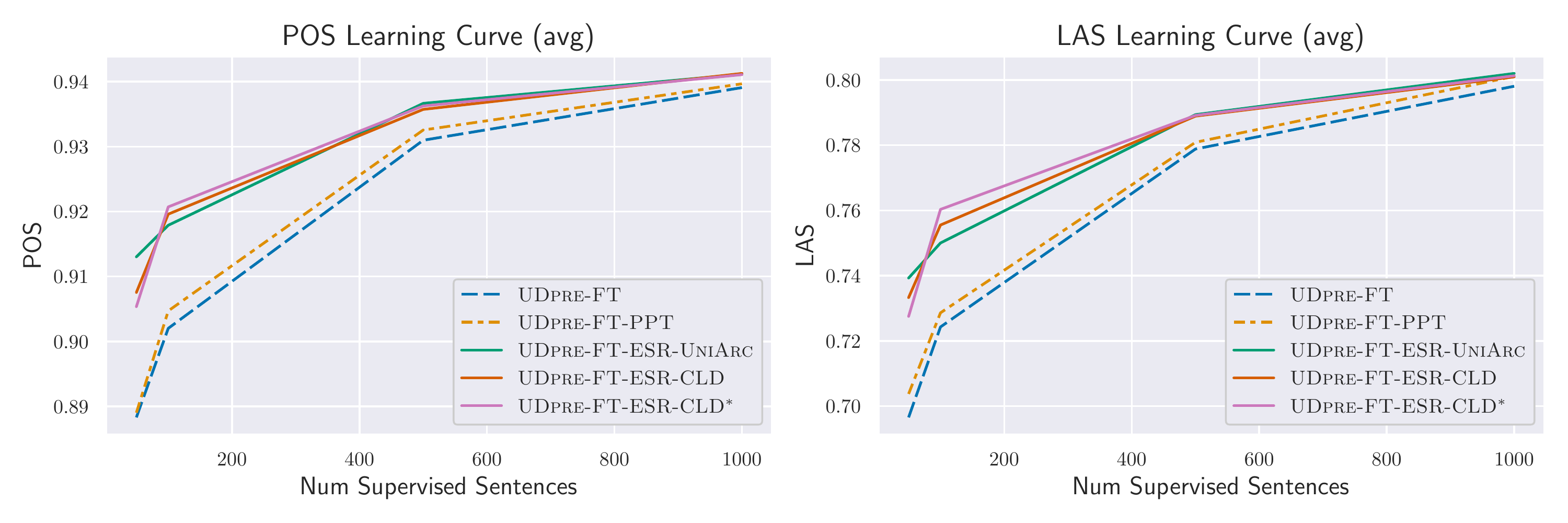} 
    \vspace{-2em}
    \caption{\textit{Multi-Source \textsc{UDpre} Transfer Learning Curves.} Baseline approaches are dotted, while \textsc{ESR} variants are solid. All curves show the average of 15 runs across 5 different languages with 3 randomly sampled labeled datasets per language. The plots indicate a significant advantage of ESR over the baselines in low-data regions.}
    \label{fig:ud-13-lc}
\end{figure*}
\begin{figure*}[t!]
    \centering
    \includegraphics[width=1.0\textwidth]{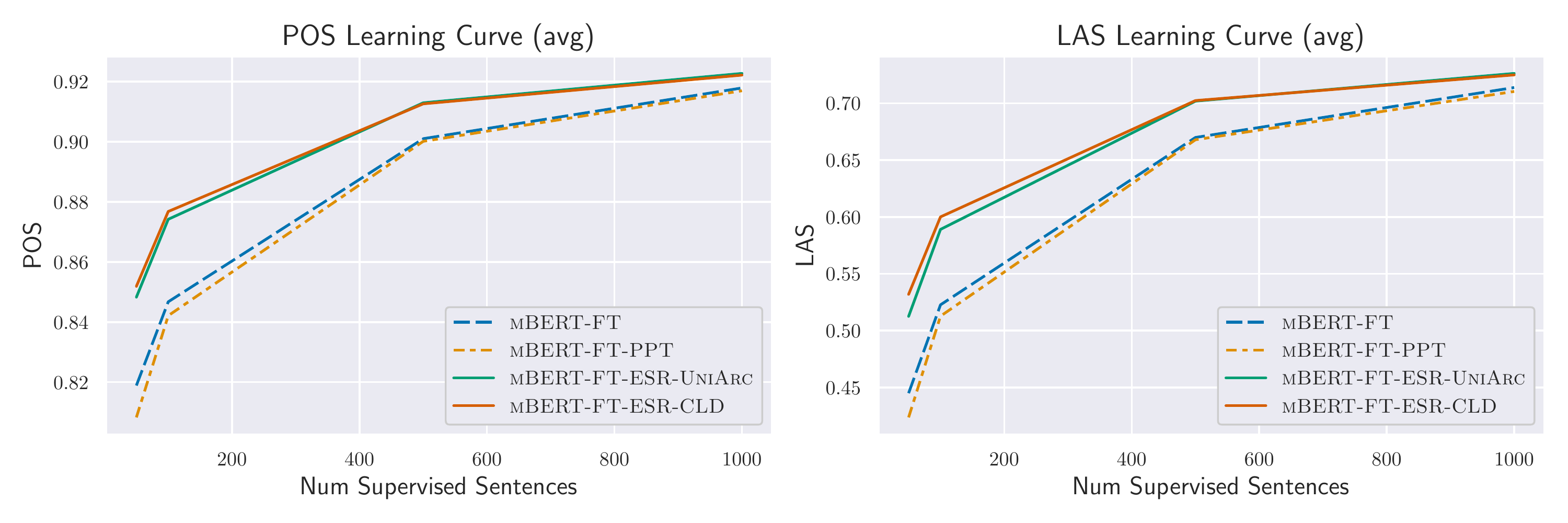} 
    \vspace{-2em}
    \caption{\textit{``From Scratch'' \textsc{mBERT} Transfer Learning Curves.} Baseline approaches are dotted, while \textsc{ESR} variants are solid. All curves show the average of 15 runs across 5 different languages with 3 randomly sampled labeled datasets per language. The plots indicate a significant advantage of ESR over the baselines in all regions.}
    \label{fig:scratch-lc}
\end{figure*}

\subsubsection{\textsc{UDpre} Transfer}
\label{s:ud-13-semi-supervised}

In this experiment, we evaluate in the multlingual transfer scenario by initializing from \textsc{UDpre}. In addition to the two chosen realistic ESR variants, we also experiment with an ``oracle'' version of \textsc{ESR-CLD}, called \textsc{ESR-CLD$^*$}, that uses target statistics estimated from the full training data. This allows us to see if small-sample estimates cause a degradation in performance compared to accurate large-sample estimates.

\noindent \textbf{Results:} Learning curves for the different approaches, averaged over all 3 runs for all 5 languages (15 total), are given in Figure~\ref{fig:ud-13-lc}. From the figure we can discern several encouraging results.

\textbf{\textsc{ESR-CLD} and \textsc{ESR-UniArc} add significant benefit to fine-tuning for small data.} Both variants significantly outperform the baselines at 50 and 100 labeled examples. For example, relative to \textsc{UDpre-FT}, the \textsc{ESR-CLD} model yielded gains of +2 POS, +3.6 LAS at 50 examples and +1.8 POS, +3.2 LAS at 100 labeled examples. At 500 and 1000 examples, however, we begin to see diminishing benefits to \textsc{ESR} on top of fine-tuning.

\textbf{\textsc{ESR-UniArc} is much more effective in conjunction with fine-tuning.} Compared to the unsupervised experiment in Section~\ref{s:stat-variants} where it ranked 25/32, the \textsc{ESR-UniArc} statistic is much more competitive with the more detailed \textsc{ESR-CLD} statistics. One potential explanation is that without labeled data (as in Section~\ref{s:stat-variants}) the \textsc{ESR-UniArc} statistic is under-specified (the 727 allowed arcs are all free to take any value), whereas the inclusion of some labeled data in this experiment fills this gap by implicitly indicating target proportions for the allowed arcs. This suggests that an approach which combines UniArc constraints with elements of self-training (like PPT) that supervise the ``free'' non-zero combinations could potentially be a useful approach to zero-shot transfer. However, we leave this to future work.

\textbf{Small-data estimates for \textsc{ESR-CLD} are as good as accurate estimates.} Comparing \textsc{ESR-CLD} to the unrealistic \textsc{ESR-CLD$^*$}, we find no significant difference between the two, indicating that, at least for the \textsc{CLD} statistic, using target estimates from small samples is as good as large-sample estimates. This may be due in part to the margin estimates $\sigma$, which are wider for the small samples and somewhat mitigate their inaccuracies.

\textbf{\textsc{PPT} adds little benefit to fine-tuning.} Relative to \textsc{UDpre-FT}, the \textsc{UDpre-FT-PPT} baseline does not yield much gain, with a maximum average improvement of +0.3 POS and +0.7 LAS over all dataset sizes. This indicates that fine-tuning and PPT-style self-training may be redundant.

\subsubsection{\textsc{mBERT} Transfer}
\label{s:scratch-semi-supervised}

In this experiment, we consider a counterfactual setting: what if the UD data was not a massively multilingual dataset where we can utilize multilingual model-transfer, and instead was an isolated dataset with no related data to transfer from? This situation reflects the more standard semi-supervised learning setting, where we are given a new task, some labeled and unlabeled data, and must build a model ``from scratch'' on that data. 

For this experiment, we repeat the learning curve setting from Section~\ref{s:ud-13-semi-supervised}, but initialize our model directly with \textbf{\textsc{mBERT}}, skipping the intermediate \textbf{\textsc{UDpre}} training.

\noindent \textbf{Results:} Learning curves for the different approaches, averaged over all 3 runs for all 5 languages (15 total), are given in Figure~\ref{fig:scratch-lc}.
The results from this experiment are encouraging; \textsc{ESR} has even greater benefits when fine-tuning directly from \textsc{mBERT} than the previous experiment, indicating that our general approach may be even more useful outside of domain-adaptation conditions.



\subsection{Low-Resource Transfer}

In previous experiments, we limited the number of evaluation treebanks to 5 to allow for variation in other dimensions (i.e., constraint types, loss types, differing amounts of labeled data). In this experiment, we expand the number of treebanks and evaluate transfer performance in a low-resource setting with only $|\D^{\text{train}}_L|=50$ labeled sentences in the target treebank, comparing \textsc{UDpre}, \textsc{UDpre-FT}, and \textsc{UDpre-FT-ESR-CLD}. As before, we subsample 3 small datasets per treebank and calculate the target statistics $t$ and margins $\sigma$ from these to make transfer results realistic.

We select evaluation treebanks according to the following criteria. For each unique language in UD v2.8 that is not one of the 13 training languages, we select the largest treebank, and keep it if has at least 250 train sentences and a development set, so that we can get reasonable variability in the subsamples. This process yields 44 diverse evaluation treebanks.

\noindent \textbf{Results:} The results of this experiment are given in Table~\ref{tbl:low-resource-results}. From the table we can see the our approach ESR (\textsc{UDpre-FT-ESR-CLD}) outperformed supervised fine-tuning (\textsc{UDpre-FT}) in many cases, often by a large margin. On average, \textsc{UDpre-FT-ESR-CLD} outperformed \textsc{UDpre-FT} by +2.6 POS and +2.3 LAS across the 44 languages. Further, \textsc{UDpre-FT-ESR-CLD} outperformed zero shot transfer, \textsc{UDpre}, by $+10.0$ POS and $+14.7$ LAS on average.

Interestingly, we found that there were several cases of large performance gains while there were no cases of large performance declines. For example, ESR improved LAS scores by $+17.3$ for Wolof,  $+16.8$ for Maltese, and $+12.5$ for Scottish Gaelic, and 9/44 languages saw LAS improvements $\geq +5.0$, while the largest decline was only $-2.5$. Additionally, ESR improved POS scores by $+20.9$ for Naija, $+11.2$ for Welsh, and 9/44 languages saw POS improvements $\geq +5.0$.

The cases of performance decline for LAS merit further analysis. Of the 20 languages with negative $\Delta$ LAS, 18 of these are modern languages spoken in continental Europe (mostly Slavic and Romance), while only 5 of the 24 languages with positive $\Delta$ LAS meet this criteria. We hypothesize that this tendency is be due to the training data used for pretraining \textsc{mBERT}, which was heavily skewed towards this category~\citep{devlin2018bert}. This suggests that ESR is particularly helpful in cases of transfer to domains that are underrepresented in pretraining.

\begin{table*}[h!]
    \newcolumntype{C}{@{\extracolsep{0.5cm}}c@{\extracolsep{0pt}}}%
    \centering
    \footnotesize
    \addtolength{\tabcolsep}{-3pt}
    \begin{tabular}{lHcccccCcccc}
    \toprule
    & & & \multicolumn{4}{c}{POS} & & \multicolumn{4}{c}{LAS} \\
    \cmidrule(lr){4-7} \cmidrule(lr){9-12}
    {Treebank} &  Lang Dist & Family &  \textsc{UDpre} &  FT &  ESR &  $\Delta$ & &  \textsc{UDpre} &  FT &  ESR &  $\Delta$\\
\midrule
Wolof-WTB                  &             0.56 &   Northern Atlantic &      40.6 &     79.5 &      \textbf{85.4} &     +5.9 &  &      12.7 &    55.9 &     \textbf{73.3} &   +17.3 \\
Maltese-MUDT               &            0.49 &  Semitic &    35.1 &     82.6 &      \textbf{91.8} &     +9.2 &   &     16.0 &    57.5 &     \textbf{74.2} &   +16.8 \\
Scottish\_Gaelic-ARCOSG     &           0.48 &  Celtic &    45.7 &     66.0 &      \textbf{75.9} &     +9.9 &   &     24.4 &    56.4 &     \textbf{68.9} &   +12.5 \\
Faroese-FarPaHC            &           0.35  & Germanic  &    74.7 &     86.2 &      \textbf{87.2} &     +1.1 &   &     43.0 &    71.4 &     \textbf{80.7} &    +9.3 \\
Gothic-PROIEL              &           0.47  &  Germanic  &   30.1 &     67.6 &      \textbf{71.7} &     +4.1 &  &      12.6 &    45.8 &     \textbf{54.6} &    +8.8 \\
Welsh-CCG                  &           0.49  &  Celtic  &    71.9 &     74.7 &      \textbf{85.8} &    +11.2 &   &     54.8 &    69.4 &     \textbf{77.6} &    +8.1 \\
Western\_Armenian-ArmTDP    &           -  &  Armenian  &   80.6 &     84.9 &      \textbf{87.1} &    +2.2 &    &    60.4 &    67.0 &     \textbf{72.7} &   +5.7 \\
Telugu-MTG                 &           0.53  & Dravidian  &    82.0 &     \textbf{81.6} &      \textbf{81.6} &    0.0 &   &     70.9 &    74.6 &     \textbf{80.1} &    +5.5 \\
Vietnamese-VTB             &          0.62   &  Viet-Muong &    67.0 &     85.6 &      \textbf{88.5} &     +2.9 &    &    46.3 &    55.3 &     \textbf{60.8} &    +5.5 \\
Turkish\_German-SAGT        &           -  &  Code Switch &    76.8 &     84.4 &      \textbf{85.8} &     +1.4 &    &    48.0 &    58.0 &     \textbf{62.1} &    +4.1 \\
Afrikaans-AfriBooms        &           0.53  & Germanic  &    90.7 &     88.0 &      \textbf{91.3} &     +3.3 &    &    62.0 &    79.4 &     \textbf{83.4} &    +3.9 \\
Hungarian-Szeged           &           0.42  &  Ugric  &   87.9 &     79.9 &      \textbf{89.7} &     +9.7 &    &    74.0 &    77.8 &     \textbf{81.7} &    +3.9 \\
Galician-CTG               &           0.37  &  Romance &    91.8 &     89.0 &      \textbf{91.2} &     +2.2 &    &    60.5 &    74.3 &     \textbf{77.8} &    +3.6 \\
Marathi-UFAL               &           0.39  & Marathi  &    71.4 &     81.1 &      \textbf{82.}3 &     +1.1 &    &    44.9 &    59.5 &     \textbf{62.5} &    +3.0 \\
Naija-NSC                  &           0.29  & Creole  &    46.5 &     68.0 &      \textbf{88.9} &    +20.9 &   &     27.9 &    71.1 &     \textbf{73.4} &    +2.3 \\
Greek-GDT                  &         0.54    & Greek  &    87.1 &     \textbf{92.8} &      92.5 &    -0.3 &    &    78.7 &    86.3 &     \textbf{88.0} &    +1.8 \\
Tamil-TTB                  &          0.54   &  Dravidian  &   72.3 &     72.4 &      \textbf{79.6} &     +7.2 &    &    46.7 &    64.9 &     \textbf{66.4} &    +1.5 \\
Indonesian-GSD             &          0.65   & Austronesian  &    82.3 &     89.8 &      \textbf{90.2} &     +0.5 &   &     58.3 &    72.9 &     \textbf{74.3} &    +1.4 \\
Uyghur-UDT                 &          0.43  &  Turkic  &   23.7 &     59.8 &      \textbf{65.5} &     +5.6 &    &    14.0 &    38.0 &     \textbf{39.2} &   +1.3 \\
Old\_French-SRCMF           &          0.28   & Romance  &    65.3 &     74.2 &      \textbf{76.2} &     +2.0 &   &     44.0 &    56.7 &     \textbf{57.8} &    +1.2 \\
Old\_Church\_Slavonic-PROIEL &          0.39   & Slavic  &    37.3 &     54.7 &      \textbf{61.0} &     +6.3 &   &     19.2 &    39.0 &     \textbf{40.1} &    +1.1 \\
Portuguese-GSD             &          0.31   & Romance  &    92.1 &     89.6 &      \textbf{92.8} &     +3.3 &   &     74.4 &    84.1 &     \textbf{84.5} &   +0.4 \\
Danish-DDT                 &         0.21    & Germanic  &    92.0 &     \textbf{92.7} &      92.1 &    -0.6 &    &    71.0 &    75.5 &     \textbf{75.7} &    +0.2 \\
Armenian-ArmTDP            &          0.47   & Armenian  &    84.7 &     \textbf{88.1} &      88.0 &    -0.1 &     &   64.1 &    69.0 &     \textbf{69.2} &    +0.1 \\
Spanish-AnCora             &          0.34   & Romance  &    94.5 &     95.2 &      \textbf{95.4} &     +0.2 &      &  77.8 &    \textbf{83.0} &     82.9 &   -0.1 \\
Catalan-AnCora             &           0.26  & Romance  &    92.9 &     94.4 &      \textbf{94.6} &     +0.3 &     &   75.8 &    \textbf{82.5} &     82.4 &   -0.1 \\
Serbian-SET                &          0.40   &  Slavic &    91.2 &     90.7 &      \textbf{93.1} &     +2.4 &    &    81.6 &    \textbf{86.5} &     86.4 &   -0.1 \\
Slovak-SNK                 &          0.39   &  Slavic  &   91.5 &     91.5 &      \textbf{92.0} &     +0.5 &      &  81.6 &    \textbf{84.0} &     83.9 &   -0.1 \\
Romanian-Nonstandard       &          0.25   &  Romance  &   79.2 &     83.3 &      \textbf{85.0} &     +1.7 &    &    54.5 &   \textbf{63.6} &     63.4 &   -0.2 \\
Polish-PDB                 &             &  Slavic &    89.7 &     90.4 &      \textbf{90.9} &     +0.5 &     &   76.0 &    \textbf{79.7} &     79.4 &   -0.3 \\
German-HDT                 &             &  Germanic  &   89.6 &     \textbf{94.4} &      94.2 &    -0.2 &     &   83.0 &    \textbf{88.2} &     87.7 &   -0.5 \\
Lithuanian-ALKSNIS         &             &  Baltic  &   87.0 &     \textbf{87.4} &      \textbf{87.4} &    0.0 &    &    65.4 &    \textbf{69.2} &     68.6 &   -0.6 \\
Latin-ITTB                 &             &  Italic  &    73.8 &     80.9 &      \textbf{81.7 }&     +0.8 &     &   51.7 &    \textbf{64.3} &     63.7 &   -0.6 \\
Bulgarian-BTB              &             &  Slavic  &    91.9 &     \textbf{94.7} &      94.6 &    -0.1 &     &   78.0 &    \textbf{84.4} &     83.7 &   -0.7 \\
Czech-PDT                  &             & Slavic  &    90.6 &     92.1 &      \textbf{92.7} &     +0.6 &     &   78.1 &    \textbf{81.9} &     81.1 &   -0.8 \\
Persian-PerDT              &             &  Iranian &    79.1 &     \textbf{91.0} &      90.8 &    -0.2 &     &   48.4 &    \textbf{74.6} &     73.7 &   -0.9 \\
Slovenian-SSJ              &             &  Slavic  &   89.2 &     90.9 &      \textbf{91.2} &     +0.3 &     &   79.6 &    \textbf{84.5} &     83.5 &   -0.9 \\
Croatian-SET               &             &  Slavic  &   91.4 &     91.7 &      \textbf{92.1} &     +0.4 &    &    80.0 &   \textbf{ 84.1} &     83.1 &   -1.0 \\
Urdu-UDTB                  &             &  Indic  &   86.9 &     \textbf{90.0} &      88.2 &    -1.8 &    &    68.7 &    \textbf{75.7} &     74.4 &   -1.3 \\
Ukrainian-IU               &             &  Slavic  &   91.5 &     92.0 &      \textbf{92.4} &     +0.3 &     &   79.6 &    \textbf{81.2} &     80.0 &   -1.3 \\
Dutch-Alpino               &             &  Germanic  &   90.0 &     \textbf{90.6} &      \textbf{90.6} &    0.0 &     &   78.9 &    \textbf{81.6} &     80.3 &   -1.3 \\
Norwegian-Bokmaal          &             &  Germanic    &  91.7 &     91.8 &      \textbf{92.1} &     +0.3 &    &    80.8 &    \textbf{82.5} &     81.0 &   -1.5 \\
Belarusian-HSE             &             &  Slavic   &  91.5 &     91.6 &      \textbf{91.9} &     +0.3 &    &    78.9 &    \textbf{79.8} &     78.1 &   -1.8 \\
Estonian-EDT               &             & Finnic  &    89.1 &     \textbf{89.6} &      89.2 &    -0.4 &    &    70.4 &    \textbf{71.4} &     68.9 &   -2.5 \\
\midrule
Average &  & & 77.3 & 84.7 & \textbf{87.3} & +2.6 & & 59.0 & 71.4 & \textbf{73.7} & +2.3\\

\bottomrule
    \end{tabular} 
    \caption{\textit{Low-Resource Semi-Supervised Transfer Results.} Transfer results for 44 unseen test languages using 50 labeled sentences in the target language, averaged over 3 subsampled datasets. ``FT'' refers to the \textsc{UDpre-FT} fine-tuning baseline, ``ESR'' refers to our \textsc{UDpre-ESR-CLD} approach, and $\Delta$ refers to the absolute difference of ESR minus FT. Best performing methods are bolded. Results are ordered from best to worst $\Delta$ LAS. } 
    \label{tbl:low-resource-results}
\end{table*}

\section{Related Work}

Related work generally falls into two categories: weak supervision and cross-lingual transfer.

\textbf{Weak Supervision:} Supervising models with signals weaker than fully labeled data has and continues to be a popular topic of interest. Current trends in weak supervision focus on generating instance-level supervision, using weak information such as: relations between multiple tasks~\citep{greenberg-etal-2018-marginal,Ratner2018SnorkelMW,Noach2019TransferLB}; labeled features~\citep{Druck2008LearningFL,Ratner2016DataPC,karamanolakis-etal-2019-leveraging}; coarse-grained labels~\citep{Angelidis2018MultipleIL,karamanolakis-etal-2019-weakly}; dictionaries and distant supervision~\citep{bellare2007learning,carlson2009learning,liu2019towards,Ustun2020UDapterLA}; or some combination of thereof~\citep{Ratner2016DataPC,karamanolakis-etal-2019-leveraging}.

In contrast, our work is more closely related to older work on population-level supervision. These techniques include Constraint-Driven Learning (CODL)~\citep{Chang2007GuidingSW}, posterior regularization (PR)~\citep{Ganchev2010PosteriorRF}, the measurements framework of \citet{Liang2009LearningFM}, and the generalized expectation criteria (GEC)~\citep{Druck2008LearningFL,Druck2009SemisupervisedLO,Mann2010GeneralizedEC}. 

Our work can be seen as an extension of GEC to more expressive expectations and to modern mini-batch SGD training. There are a two more recent works that touch on these ideas, but both have significant downsides compared to our approach. \citet{Meng2019TargetLC} use a PR approach inspired by \citet{Ganchev2013CrossLingualDL} for cross-lingual parsing, but must use very simple constraints and require a slow inference procedure that can only be used at test time. \citet{Noach2019TransferLB} utilize GEC with minibatch training, but focus on using related tasks for computing simpler constraints and do not adapt their targets to small batch sizes.

\textbf{Cross-Lingual Transfer:} Earlier trends in cross-lingual transfer for parsing used dexlexicalization~\citep{Zeman2008CrossLanguagePA,McDonald2011MultiSourceTO,Tckstrm2013TargetLA} and then aligned multilingual word vector-based approaches~\citep{Guo2015CrosslingualDP,Ammar2016ManyLO,Rasooli2017CrossLingualST,ahmad-etal-2019-cross}.
With the rapid rise of language-model pretraining \citep{Peters2018DeepCW,devlin2018bert,liu2019roberta}, recent research has focused on multilingual PLMs and multitask fine-tuning to achieve generalization in transfer.  \citet{wu-dredze-2019-beto} showed that a multilingual PLM afforded surprisingly effective cross-lingual transfer using only English as the fine-tuning language. 
\citet{Kondratyuk201975L1} extended this approach by fine-tuning a PLM on the concatenation of all treebanks. \citet{Tran2019ZeroshotDP}, however, show that transfer to distant languages benefit less.  

Other recent successes have been found with linguistic side-information~\citep{Meng2019TargetLC,Ustun2020UDapterLA}, careful methodology for source-treebank selection~\citep{Tiedemann2016SyntheticTF,Tran2019ZeroshotDP,lin-etal-2019-choosing,Glavas2021ClimbingTT}, self-training~\citep{Kurniawan2021PPTPP}, and paired bilingual text for annotation projection~\citep{rasooli-tetrault-2015,Rasooli2019LowResourceST,Liu2020CrossLingualDP,Shi2021SubstructureDP}. 

\section{Conclusion}
\label{s:conclusion}

We have presented Expected Statistic Regularization, a general approach to weak supervision for structured prediction, and studied it in the context of modern cross-lingual multi-task syntactic parsing. We evaluated a wide range of expressive structural statistics in idealized and realistic transfer scenarios and have shown that the proposed approach is effective and complementary to the state-of-the-art model-transfer approaches.



\section*{Acknowledgements}

We would like to thank Chris Kedzie, Giannis Karamanolakis, and the reviewers for helpful conversations and feedback.

\bibliography{tacl2021}
\bibliographystyle{acl_natbib}

\end{document}